\documentclass[12pt]{article}
\pdfoutput=1

\usepackage[preprint]{NECO_mod}

\usepackage{setspace,caption}

\usepackage[subrefformat=parens]{subcaption}
\usepackage{hyperref}
\usepackage{url}
\usepackage{color}
\usepackage{amssymb, latexsym, mathtools}
\newcommand{\com}[1]{\textcolor{blue}{#1}}
\renewcommand{\com}[1]{\textcolor{black}{#1}}
\newcommand{\fixme}[1]{{\color{red}{#1}}}
\renewcommand{\fixme}[1]{{\color{black}{#1}}}

\usepackage{amsmath,amsfonts,bm}

\usepackage{graphicx}

\def\eqref#1{equation~\ref{#1}}

\def\1{\bm{1}}

\def\vh{{\bm{h}}}

\def\vv{{\bm{v}}}

\DeclareMathAlphabet{\mathsfit}{\encodingdefault}{\sfdefault}{m}{sl}
\SetMathAlphabet{\mathsfit}{bold}{\encodingdefault}{\sfdefault}{bx}{n}

\newcommand{\R}{\mathbb{R}}

\newcommand{\softmax}{\mathrm{softmax}}

\newcount\K 
\def\blah{
    \K=0 \loop\ifnum\K<20 
    {\color[rgb]{0.8, 0.8, 0.8} blah blah blah blah blah blah blah blah blah blah} 
    \advance\K by1\repeat 
}

\usepackage{mleftright}
\mleftright

\newcommand{\kh}{generalized Hopfield network} 
\newcommand{\abm}{AttnBM} 

\newcommand{\xvis}{v}

\newcommand{\Z}{\mathbb{Z}}

\begin{document}
\hspace{13.9cm}1
\ \vspace{20mm}\\

{\LARGE Attention in a family of Boltzmann machines emerging from modern Hopfield networks}

\ \\
{\bf \large Toshihiro Ota$^{\displaystyle 1, \displaystyle 2}$ \& Ryo Karakida$^{\displaystyle 3}$}\\
{$^{\displaystyle 1}$CyberAgent, Inc.}\\
{$^{\displaystyle 2}$RIKEN.}\\  
{$^{\displaystyle 3}$AIST.}
%

{\bf Keywords:} Hopfield networks, Boltzmann machines

\thispagestyle{empty}
\markboth{}{NC instructions}
\ \vspace{-0mm}\\

\begin{center} {\bf Abstract} \end{center}

Hopfield networks and Boltzmann machines (BMs) are fundamental energy-based neural network models.
Recent studies on modern Hopfield networks have broaden the class of energy functions and led to
a unified perspective on general Hopfield networks including an attention module.
In this letter, we consider the BM counterparts of modern Hopfield networks using the associated energy functions,
and study their salient properties from a trainability perspective.
In particular, the energy function corresponding to the attention module naturally introduces a novel BM,
which we refer to as the attentional BM (\abm).
We verify that \abm ~has a tractable likelihood function and gradient for certain special cases and is easy to train.
Moreover, we reveal the hidden connections between \abm ~and some single-layer models,
namely the Gaussian--Bernoulli restricted BM and the denoising autoencoder with softmax units coming from denoising score matching.
We also investigate BMs introduced by other energy functions and
show that the energy function of dense associative memory models gives BMs belonging to Exponential Family Harmoniums.

\section{Introduction}

The Hopfield network is a classical associative memory model of neural networks \citep{hopfield82,hopfield84}.
In this network, stored memories and their retrieval are well described by the attractors of an energy function and the dynamics converging to them.
In practice, however, the classical Hopfield network is known to suffer from a limited memory capacity.
To address this issue, models with significantly increased memory capacities have recently been proposed
\citep{NIPS2016_eaae339c,demircigil2017model},
but they are supposed to have many-body interactions among neurons, which is biologically implausible in the brain.

Close to the time of the proposals of these modern Hopfield networks,
\citet{ramsauer2021hopfield} found that each of the attention modules in the Transformer \citep{NIPS2017_3f5ee243}
can be essentially identified with the process of the update rule of a certain continuous Hopfield network,
and their algorithm has been applied to some tasks with success \citep{NEURIPS2020_da4902cb}.
Based on their heuristic findings and earlier works, Krotov and Hopfield developed a more general associative memory model, which we refer to as the \kh,
and the model consists of visible and hidden neurons with only two-body interactions between them \citep{krotov2021large}.
In their paper, Krotov and Hopfield demonstrated that many of the energy-based associative memory models in the literature
are derived from a single Hopfield-type network equipped with Lagrangians defining the systems;
\fixme{%
    They presented three concrete examples referred to as models A, B and C by specifying Lagrangians.
    The model A realizes the series of dense associative memory models \citep{NIPS2016_eaae339c,demircigil2017model}
    including the classical Hopfield network, and the model B reproduces the model of Ramsauer et al.
    Also, the model C provides a new energy-based associative memory model,
    which may have a relation to MLP-Mixer \citep{tolstikhin2021mlp} as pointed out by \citep{tang2021remark}.
    The \kh ~potentially can further produce new energy-based associative memory models by replacing the Lagrangians,
    and in this sense, this neural network generates a family of Hopfield networks. 
}%

As discussed by Krotov and Hopfield, the \kh ~gives us a unified perspective for a family of Hopfield-type networks
with energy functions through the corresponding Lagrangians. 
This fact motivates us to consider the Boltzmann machine (BM) counterpart of their model by introducing the stochastic nature
and to classify a family of BMs based on Lagrangians and energy functions.
As studies on Hopfield networks lack the perspective on learning,
the BM counterpart is expected to provide complementary insights into the model trainability.

\fixme{%
In this letter, we propose the BM counterparts of the series of modern Hopfield networks from the viewpoint of the \kh,  
and study some theoretical properties of the model-A, -B, and -C BMs.
We mainly focus on the BM counterpart of the model B, which we refer to as the {\it attentional Boltzmann machine} (\abm).
We can regard \abm ~as a natural extension of the BM to an attention module.
Interestingly, owing to its salient property, \abm ~is exactly solvable in the sense that the likelihood function of the model is tractable for certain special cases.
In more detail, our contributions are listed as follows:
\begin{enumerate}
    \item We provide a general formulation of \abm, the BM counterpart of the model B, at finite temperature in Sec.~\ref{sec:modelB}.
    We show that the model is exactly solvable for some special temperatures. This can be regarded as a generalization of \citep{bal2020energyattention}
    which discussed the possibility of model-B BM under a certain temperature.
    We also clarify that the likelihood functions of general temperatures are bounded by the solvable one.
    \item We explore the basic properties of \abm.
    From theoretical perspective, we reveal the hidden connections between \abm ~and some single-layer models.
    \abm ~can be regarded as a truncated approximation of the Gaussian--Bernoulli restricted BM (G--B RBM) (Sec.~\ref{sec:sec3_3}).
    Moreover, if we consider denoising score matching for \abm, the objective function becomes equivalent to the denoising autoencoder with softmax units (Sec.~\ref{sec:sec3_4}).
    Although our main purpose is to characterize the theoretical properties of \abm,
    we also provide the first empirical demonstration on training \abm ~by the exact gradient descent of the log-likelihood in Sec.~\ref{sec:experiments}.%
    \footnote{%
        The code is available at \url{https://github.com/Toshihiro-Ota/AttnBM}.
    }%
    \item We investigate the mathematical properties of the model-A and -C BMs in Sec.~\ref{sec:modelAC}. In particular,
    we find that the energy function of dense associative memory models gives BMs belonging to Exponential Family Harmoniums \citep{welling2004exponential},
    which suggests that the model-A BMs can easily be trained by contrastive divergence.
\end{enumerate}
}%

\section{Background}

To describe notations, we here provide the derivation of the attention mechanism in Transformers \citep{NIPS2017_3f5ee243}
from a continuous Hopfield network.

\subsection{Overview of generalized Hopfield network}  
\label{sec:Overview}

Let us first briefly review the \kh ~proposed in \citep{krotov2021large}.
In this system, the dynamical variables are composed of $N_v$ visible neurons and $N_h$ hidden neurons both continuous,
\begin{align}
    \vv: \, &U (\subset \R) \to \R^{N_v},  \\
    \vh: \, &U (\subset \R) \to \R^{N_h}.
\end{align}
The interaction matrices between them,
\begin{align}
    \Xi \in \R^{N_h \times N_v},
    \quad
    \tilde{\Xi} \in \R^{N_v \times N_h},
\end{align}
are basically supposed to be symmetric: $\tilde{\Xi} = \Xi^{\top}$. 
%
%
With the relaxing time constants of the two groups of neurons $\tau_v$ and $\tau_h$,
the system is described by the following differential equations,
\begin{align}
    \tau_v \frac{d v_{i}(t)}{d t}
        &=\sum_{\mu=1}^{N_{h}} \xi_{i \mu} f_{\mu}(\vh(t)) - v_{i}(t),  \\
    \tau_h \frac{d h_\mu(t)}{d t}
        &=\sum_{i=1}^{N_{v}} \xi_{\mu i} g_{i}(\vv(t)) - h_{\mu}(t),  \label{eq:heq}
\end{align}
where the argument $t$ can be thought of as ``time'', and the activation functions $f$ and $g$ are determined through Lagrangians
$L_h: \R^{N_h}\to \R$ and $L_v: \R^{N_v}\to \R$, such that
\begin{align}
    f_\mu(\vh) = \frac{\partial L_{h}(\vh)}{\partial h_{\mu}},
    \quad
    g_i(v) = \frac{\partial L_{v}(\vv)}{\partial v_{i}}.
    \label{eq6:1126}
\end{align}


The canonical energy function for this system is given as
\begin{align}
    E(\vv, \vh) = \sum_{i=1}^{N_v} v_{i} g_{i}(\vv) - L_{v}(\vv) 
        + \sum_{\mu=1}^{N_h} h_{\mu} f_{\mu}(\vh) - L_{h}(\vh)
        - \sum_{\mu, i} f_{\mu} \xi_{\mu i} g_{i}.
        \label{eq:eng}
\end{align}
One can easily find that this energy function monotonically decreases along the trajectory of the dynamical equations,
provided that the Hessians of the Lagrangians are positive semi-definite:
\begin{align}
    \frac{d E(\vv(t), \vh(t))}{d t} \leq 0.
\end{align}

Suppose we have a fixed interaction matrix $\xi_{\mu i}$,
then the system is defined by the choice of Lagrangians $L_h$ and $L_v$.
In this sense, this neural network generates a family of Hopfield networks by choosing the corresponding Lagrangians.
Krotov and Hopfield presented three concrete examples, models A, B and C by specifying Lagrangians.
In the next subsection we review the model B, and the others will be discussed in Sec.~\ref{sec:modelAC}.

\subsection{Attention mechanism}
\label{sec:AttMec}

Krotov and Hopfield demonstrated that the specific choice of Lagrangians called model B in their paper \citep{krotov2021large}
essentially reproduces the attention mechanism in Transformers \citep{NIPS2017_3f5ee243,ramsauer2021hopfield}.
This model is given by the following Lagrangians:
\begin{align}
    L_h(\vh) = \log\sum_{\mu} e^{h_\mu},
    \quad
    L_v(\vv) = \frac12 \sum_i v_i^2. \label{eq:Lattn}
\end{align}
For these Lagrangians, the activation functions are
\begin{align}
    f_\mu(\vh)
        &= \frac{\partial L_h}{\partial h_\mu}
        = \frac{e^{h_\mu}}{\sum_\nu e^{h_\nu}}
        = \mathrm{softmax}(h_\mu),  \\
    g_i(\vv)
        &= \frac{\partial L_v}{\partial v_i}
        = v_i.
\end{align}

Now we assume the adiabatic limit, $\tau_v \gg \tau_h$,
which means that the dynamics of the hidden neurons is much faster than that of the visible neurons,
i.e., we can take $\tau_h \to 0$:
\begin{align}
    \text{Eq.~(\ref{eq:heq})}
    \quad \rightsquigarrow \quad
    h_{\mu}(t) = \sum_{i=1}^{N_{v}} \xi_{\mu i} v_i(t).
    \label{eq:hadia}
\end{align}
By substituting the above expression into the other dynamical equation
and discretizing it by taking $\Delta t = \tau_v$,
then we obtain the update rule for the visible neurons,
\begin{align}
    v_i(t+1) = \sum_{\mu=1}^{N_{h}} \xi_{i \mu} \, \mathrm{softmax}\left(\sum_{j=1}^{N_{v}} \xi_{\mu j} v_j(t)\right).
    \label{eq:attn}
\end{align}
The energy function is also determined by the Lagrangians:
\begin{align}
    E_{\mathrm{B}} = \frac{1}{2}\sum_{i=1}^{N_v} v_i^2 - \log \left(\sum_{\mu} \exp \left(\sum_{i} \xi_{\mu i} v_{i}\right)\right), \label{eq13:1129}
\end{align}
where the subscript stands for the model B.
This update rule and the energy function coincide (up to some constants)
with one of the modern continuous Hopfield networks discussed in \citep{ramsauer2021hopfield},
in which the update rule of the neurons is identified with the self-attention modules in Transformers.

\section{Attentional Boltzmann Machine}
\label{sec:attnBM}

We now move on to the stochastic description of the \kh,
which yields a family of Boltzmann machines by corresponding energy functions determined through Lagrangians.
In this section, we focus on the model-B case reviewed in the previous section,
in which the update rule of the visible neurons is identified with the attention mechanism.
We refer to the model-B BM as the attentional BM (\abm).

\subsection{Stochastic description of Model B}
\label{sec:modelB}

We begin with the reformulation of Model B reviewed in Sec.~\ref{sec:AttMec} for the transparency of the discussion.
Let us consider a set of $N$ neurons which is regarded as a continuous neuron configuration on $\Z_N$:%
\footnote{This type of setup is known as a continuous spin system on a lattice $\Z_N$ in the physics literature.}
\begin{align}
    \vv = (v_1, \dots, v_N)^\top \in \R^N.  
\end{align}
A configuration $\vv$ will be treated as a random variable in the stochastic description below.

Now let us suppose we have a set of $p$ (fixed) configurations
\begin{align}
    {\bm \xi}_1, \dots, {\bm \xi}_p \in \R^N  
\end{align}
which we would like to encode in a system as memories.  
We then define an energy function of a modern Hopfield network,  
$E_{\mathrm{B}} : \R^N \to \R$,
such that it takes the same form as in Sec.~\ref{sec:AttMec},
\begin{align}
    E_{\mathrm{B}}(\vv)
        &= \frac{1}{2} \| \vv \|^2
        - \log \left( \sum_{\mu=1}^{p} \exp ({\bm \xi}^\top_\mu \vv) \right),  \label{eq18:1126} \\
    \Xi &= ({\bm \xi}_1, \dots, {\bm \xi}_p)^{\top} \in \R^{p\times N}.
\end{align}

A stochastic description of this system is introduced by the Gibbs--Boltzmann distribution with the above energy function,
and it defines a Boltzmann machine, which we refer to as the \abm:
\begin{align}
    p_{\text{\abm}}(\vv; \Xi, \beta)
    = \frac{e^{-\beta E_{\mathrm{B}}(\vv)}}{Z(E_{\mathrm{B}}; \Xi, \beta)},
    \label{eq:pmhn}
\end{align}
where $\beta(>0)$ is an inverse temperature parameter.
The denominator is the partition function of this system,
\begin{align}
    Z(E_{\mathrm{B}}; \Xi, \beta)
    &:= \int d\vv \, e^{-\beta E_{\mathrm{B}}(\vv)} \nonumber \\
    &=  \int d\vv \, e^{-\frac{\beta}{2} \|\vv\|^2}
        \exp\left( \beta \log \sum_{\mu=1}^{p} \exp ({\bm \xi}^\top_\mu \vv) \right),
\end{align}
where $\int d\vv = \int_{\R^N} \prod_{i=1}^N d\xvis_i$.
Although this integral is generically intractable for an arbitrary $\beta$, a salient property of this system is that
it becomes tractable for a special case; that is, if we restrict $\beta$ to be positive integers.
In the following, we first introduce the special case of $\beta=1$, followed by the general cases.
We clarify that $\beta=1$ has a special meaning for the general $\beta$.

\noindent
{\bf (i) Case of $\beta=1$.}
One can easily find that the partition function reduces to a simple Gaussian integral
and reads $Z(E_{\mathrm{B}}; \Xi, 1)=(2\pi)^{N/2}\sum_\mu \exp(\|{\bm \xi}_\mu\|^2/2)$.
\fixme{The model distribution becomes the following mixture of Gaussian distributions:}
\begin{align}
    p_{\text{\abm}}(\vv)
        &= \sum_{\mu=1}^p e^{-\frac{\beta}{2}\|\vv\|^2 +{\bm \xi}_{\mu}^\top \vv} / Z(E_{\mathrm{B}}; \Xi, 1) \\
        &=(2\pi)^{-N/2} \sum_{\mu=1}^p
            \softmax\big( \|{\bm \xi}_{\mu}\|^2 /2 \big)
            \exp \left(-\frac{\|\vv-{\bm \xi}_\mu\|^2}{2}\right), \label{eq22:revise}
\end{align}
where $\|{\bm \xi}_{\mu}\|^2 = \sum_j \xi_{\mu j}^2$.
Roughly speaking, when the weight vector ${\bm \xi}_\mu$ has a large norm, this unit contributes to gathering more attention.
Note that we assume no bias term in the energy function in Eq.~(\ref{eq18:1126}) for simplicity.
It is straightforward to include bias terms and shift the origins of the softmax and Gaussian distributions.
\fixme{%
Focusing on the case of $\beta=1$, \citet{bal2020energyattention} reported the tractability of $Z(E_{\mathrm{B}}; \Xi, 1)$
and the Gaussian mixture property of $p_{\text{\abm}}$, although the expression with softmax functions (\ref{eq22:revise}) was not explicitly shown.
As we reveal in Section \ref{sec:sec3_3}, 
this Gaussian mixture property has a certain connection to the G--B RBM.
Furthermore, one can find a more explicit relation between \abm ~and softmax units as is shown in Section \ref{sec:sec3_4}.
}%

\noindent
{\bf (ii) Case of positive integer $\beta$.}
The expression of the partition function simplifies to
\begin{align}
    Z(E_{\mathrm{B}}; \Xi, \beta)
    = \sum_{\mu_1,\dots,\mu_\beta} \int d\vv \, e^{-\frac{\beta}{2}\|\vv\|^2 + \sum_{k=1}^{\beta}{\bm \xi}_{\mu_k}^\top \vv}.
\end{align}
By completing the square and performing the Gaussian integral, we obtain
\begin{align}
    Z(E_{\mathrm{B}}; \Xi, \beta)
    &= \sum_{\mu_1,\dots,\mu_\beta} \int d\vv \exp\left( -\frac{\beta}{2}\|\vv - \frac{1}{\beta} \sum_k {\bm \xi}_{\mu_k }\|^2 + \frac{1}{2\beta} \| \sum_k{\bm \xi}_{\mu_k} \|^2 \right) \nonumber \\
    &= \left(\frac{2\pi}{\beta}\right)^{N/2} \sum_{\mu_1,\dots,\mu_\beta} \exp\left( \frac{1}{2\beta} \| \sum_k{\bm \xi}_{\mu_k} \|^2 \right).
    \label{eq:Zmhn}
\end{align}
Therefore, in this case, \abm ~is explicitly written by
\begin{multline}
p_{\text{\abm}}(\vv)
    =\sum_{\mu_1,\dots,\mu_\beta} e^{-\frac{\beta}{2}\|\vv\|^2 + \sum_{k=1}^{\beta}{\bm \xi}_{\mu_k}^\top \vv} / Z(E_{\mathrm{B}}; \Xi, \beta) \\  
    =\left(\frac{2\pi}{\beta}\right)^{-N/2} \sum_{\mu_1,\dots,\mu_\beta}
        \softmax\left(\frac{1}{2\beta} \| \sum_k{\bm \xi}_{\mu_k} \|^2  \right)  
        \exp \left( -\frac{\beta}{2} \|\vv - \frac{1}{\beta} \sum_k{\bm \xi}_{\mu_k } \|^2 \right),
    \label{eq:beta_int}
\end{multline}
where the softmax in the last expression is taken along all the $\mu_1,\dots,\mu_\beta$ directions.
This indicates that generically \abm ~(for $\beta$ being a positive integer) is a Gaussian mixture model (GMM)
with $p^\beta$ Gaussians and specific mixture weights given by the attention.
\com{%
    Note that in order to construct this GMM with $p^\beta$ Gaussians, it is enough to consider the case of $\beta=1$ with a sufficiently large number of hidden units $p$.
    Let us write $p_{\text{AttnBM}}$ with $p$ and $\beta$ by $p_{\text{AttnBM}}[p, \beta]$.
    We can easily find that $p_{\text{AttnBM}}[p=\bar{p}, \beta=\bar{\beta} \text{ (positive integer)}]$ is included in $p_{\text{AttnBM}}[p=\bar{p}^{\bar{\beta}}, \beta=1]$.
    From this fact, we will always set $\beta$ to be one in the following subsections.
    }%

\noindent
{\bf (iii) Case of real-valued $\beta$.}
Finally, let us make a comment on general $\beta(>1)$.
From Jensen's inequality, we have
\begin{align}
    \bigg( \sum_\mu e^{({\bm \xi}^\top_\mu \vv)} \bigg)^\beta
    \leq p^{\beta-1} \sum_\mu e^{\beta ({\bm \xi}^\top_\mu \vv)}.
\end{align}
Thus, we obtain
\begin{align}
    \log Z(E_{\mathrm{B}}; \Xi, \beta)
    &\leq \log \int d\vv \, e^{-\frac{\beta}{2}\|\vv\|^2} p^{\beta-1} \sum_\mu e^{\beta ({\bm \xi}^\top_\mu \vv)}  \nonumber \\
    &= \log p^{\beta-1} \left(\frac{2\pi}{\beta}\right)^{N/2} \sum_\mu \exp\left( \frac{\beta}{2} \| {\bm \xi}_{\mu} \|^2
    \right) \\
    &=\log Z(E_{\mathrm{B}};\sqrt{\beta}\Xi,1) + const.
\end{align}
As the negative log-likelihood function is given by $-\log p_{\text{\abm}} = \beta E_{\mathrm{B}} +\log Z$,
the above inequality implies that the partition function of $\beta=1$ works as an upper bound of the real-valued case.

\subsection{Gradient of likelihood function}
\label{sec:sec3_2}

The maximum-likelihood estimation is one of the most common approaches for training stochastic models.
To optimize the likelihood function by gradient descent, we need to evaluate
$-\nabla_{\Xi} \log p_{\text{\abm}}= \beta \nabla_{\Xi} E_{\mathrm{B}} + \nabla_{\Xi} \log Z $.
The second term is referred to as the ``negative phase'' or ``correlation function'' and usually requires a substantial computational cost.
In our case, it is given by
\begin{align}
    \frac{\partial}{\partial \xi_{\mu i}} \log Z(\Xi,\beta)
        &= \frac1Z \frac{\partial}{\partial \xi_{\mu i}} Z(\Xi,\beta)  \nonumber \\
        &= \frac{\beta}{Z} \int d\vv \, \frac{\xvis_i \exp ({\bm \xi}^\top_\mu \vv)}{\sum_\nu \exp ({\bm \xi}^\top_\nu \vv)} e^{-\beta E_{\mathrm{B}}(\vv)}  \nonumber \\
        &= \beta \left\langle \xvis_i \, \softmax ({\bm \xi}^\top_\mu \vv) \right\rangle_{\text{\abm}},
\end{align}
where $Z=Z(\Xi,\beta)=Z(E_{\mathrm{B}}; \Xi, \beta)$, and $\left\langle \cdot \right\rangle_{\text{\abm}}$ denotes
the expectation value with respect to $p_{\text{\abm}}$ in Eq.~(\ref{eq:pmhn}).
For $\beta$ being a positive integer, we can again exactly compute it from Eq.~(\ref{eq:Zmhn}).
In particular, for $\beta=1$ it is simply given as
\begin{align}
    \left\langle \xvis_i \, \softmax ({\bm \xi}^\top_\mu \vv) \right\rangle_{\text{\abm}}\mid_{\beta=1}
        &= \frac{\partial}{\partial \xi_{\mu i}} \log\left( (2\pi)^{N/2} \sum_{\nu=1}^p \exp\left({\sum_{j=1}^N \xi_{\nu j}^2/2}\right) \right)  \nonumber \\
        &= \xi_{\mu i} \, \softmax \left( \|{\bm \xi}_\mu\|^2 / 2 \right). \label{eq27:1129}
\end{align}

It is noteworthy that there have been only a few examples of Boltzmann machines whose normalization factor and gradient are computationally tractable.
The Gaussian--Gaussian RBM is tractable and has been used to theoretically analyze contrastive divergence learning \citep{williams2002analysis}.
We can explicitly analyze all the stable fixed points although the model has a limited representational power,
and can extract only principal components from data \citep{karakida2016dynamical}.
If we restrict the weight matrix to be orthogonal and the number of hidden units undercomplete (i.e., $N_h \leq N_v$),
the G--B RBM also has a tractable normalization factor and gradient \citep{karakida2016GB}.
In contrast, the tractability of \abm ~requires no condition on the weight and allows the overcomplete case.
Thus, \abm ~is a new and valuable example of a tractable BM that we can easily compute the exact gradient without any sampling methods.

\subsection{Connection to Gaussian--Bernoulli RBM}
\label{sec:sec3_3}

Hereafter, we rewrite $N$ to $N_v$ and $p$ to $N_h$ in \abm ~to make the connection with single-layer models more clear.
As \abm ~is a GMM, it is also expected to be related to the G--B RBM \citep{Hinton06,liao2022gaussian},
which is another GMM. The G--B RBM is defined by
\begin{align}
    p_{\mathrm{GB}}(\vv,\vh)
        &= e^{ -E_{\mathrm{GB}}(\vv,\vh) } /Z, \label{eq28:1127} \\
    E_{\mathrm{GB}}(\vv,\vh)
        &= \frac{1}{2\sigma^2}\|\vv\|^2 - \vh^\top W \vv,
\end{align}
where we set $h_\mu\in\{0,1\}$ $(\mu=1,\dots,N_h)$.
For simplicity, we assume that visible units have a non-zero mean and constant variance.
We have
\begin{equation}
    p_{\mathrm{GB}}(\vv)
        = \prod_\mu \left[1+ e^{w_\mu \vv}\right] e^{ -\frac{1}{2} \|\vv\|^2} /Z,
\end{equation}
where $w_\mu$ is the $\mu$-th row of $W$. By expanding the product part, we obtain
\begin{equation}
    p_{\mathrm{GB}}(\vv)
        = \Big[ 1+ \sum_\mu e^{w_\mu \vv} + \underbrace{\sum_{\mu \neq \mu'} e^{(w_\mu + w_{\mu'}) \vv} + \dots + e^{(\sum_\mu w_\mu)\vv}}_{=:H(\vv;w)} \Big]
        e^{ -\frac{1}{2} \|\vv\|^2}/Z.
\end{equation}
The distribution is a mixture of $2^{N_h}$ Gaussian distributions whose location is spanned by a combination of $N_h$ weight vectors.
We can see that if we punctuate the expansion in $p_{\mathrm{GB}}$ and neglect the higher-order terms $H(\vv;w)$,
this is equivalent to $p_{\text{\abm}}(\vv)$ with $N_h+1$ hidden units including $w_0=0$. Instead of $w_0$, we may set a bias term.
A typical G--B RBM is a GMM with an exponential number of Gaussian distributions, which makes the computation of the partition function hard in practice.
In contrast, \abm ~requires only $N_h(+1)$ Gaussians.
Although this may result in the expressive power being smaller than that of the G--B RBM,
it is still interesting in its own right that \abm, the new example of a Boltzmann machine with the attension mechanism,
has a connection to the seemingly different RBM.

\subsection{Connection to denoising autoencoder}
\label{sec:sec3_4}

The maximum-likelihood estimation is not the only way to train BMs.
Score matching and its variants are common for training BMs with continuous visible units \citep{swersky2011autoencoders,karakida2016adaptive}.
Here, let us consider training with denoising score matching.
\abm ~then corresponds to a denoising autoencoder with softmax units as follows.

As revealed by \citep{vincent2011connection}, the training of the G--B RBM by denoising score matching is
equivalent to the training of a denoising autoencoder with sigmoid hidden units.
The objective function of denoising score matching is known as
\begin{equation}
    J_{\text{DSM}}(W)
        := \left\langle \left\| \Psi(\tilde{\vv};W) - \frac{\partial \log q(\tilde{\vv}|\vv)}{\partial \tilde{\vv}}  \right\|^2 \right\rangle_{q(\tilde{\vv}|\vv)q_0(\vv)},
\end{equation}
where the score function is defined as $\Psi(\vv;W)=\partial \log p(\vv)/\partial \vv$,
$q_0$ denotes the empirical input distribution, and $q(\tilde{\vv}|\vv)= \exp(-\|\tilde{\vv}-\vv\|^2/2)/(2\pi)^{N_v/2}$.
For the G--B RBM (\ref{eq28:1127}), we have
\begin{equation}
    J_{\text{DSM on GB}}(W)
        = \left\langle \left\| W^\top \mathrm{sigmoid}(W\tilde{\vv}) - \vv  \right\|^2 \right\rangle_{q(\tilde{\vv}|\vv)q_0(\vv)},
\end{equation}
which is equivalent to the denoising autoencoder.
We can consider the case of our \abm ~similarly. Following a straightforward calculation,
we have $\Psi(\vv; \Xi)= {\Xi}^\top \softmax(\Xi \vv) -\vv$ and $\partial \log q(\tilde{\vv}|\vv)/\partial \tilde{\vv}=\vv-\tilde{\vv}$.
Therefore, we obtain
\begin{equation}
    J_{\text{DSM on \abm}}(\Xi) = \left \langle \left \| \Xi^\top \softmax(\Xi \tilde{\vv}) - \vv  \right \|^2 \right \rangle_{q(\tilde{\vv}|\vv)q_0(\vv)}.
\end{equation}
The denoising score matching of \abm ~is equivalent to the denoising autoencoder with hidden softmax units.
We can say that \abm ~is hidden behind the unsupervised learning of the attention module via Gaussian denoising.

\subsection{Numerical experiments}
\label{sec:experiments}


Although the primary purpose of this section is to explore the mathematical features of \abm ~emerging from the model B,
it will be informative to briefly show several toy experiments on training.
\fixme{%
In the following, we demonstrate image reconstruction and the visualization of the receptive fields of trained \abm.
We show two typical behaviors: one is the memorization of training samples for a small sample size,
and the other is the extraction of localized features for a large sample size.
}%


\begin{figure}[t]
    \centering
    \includegraphics[width=0.7\textwidth]{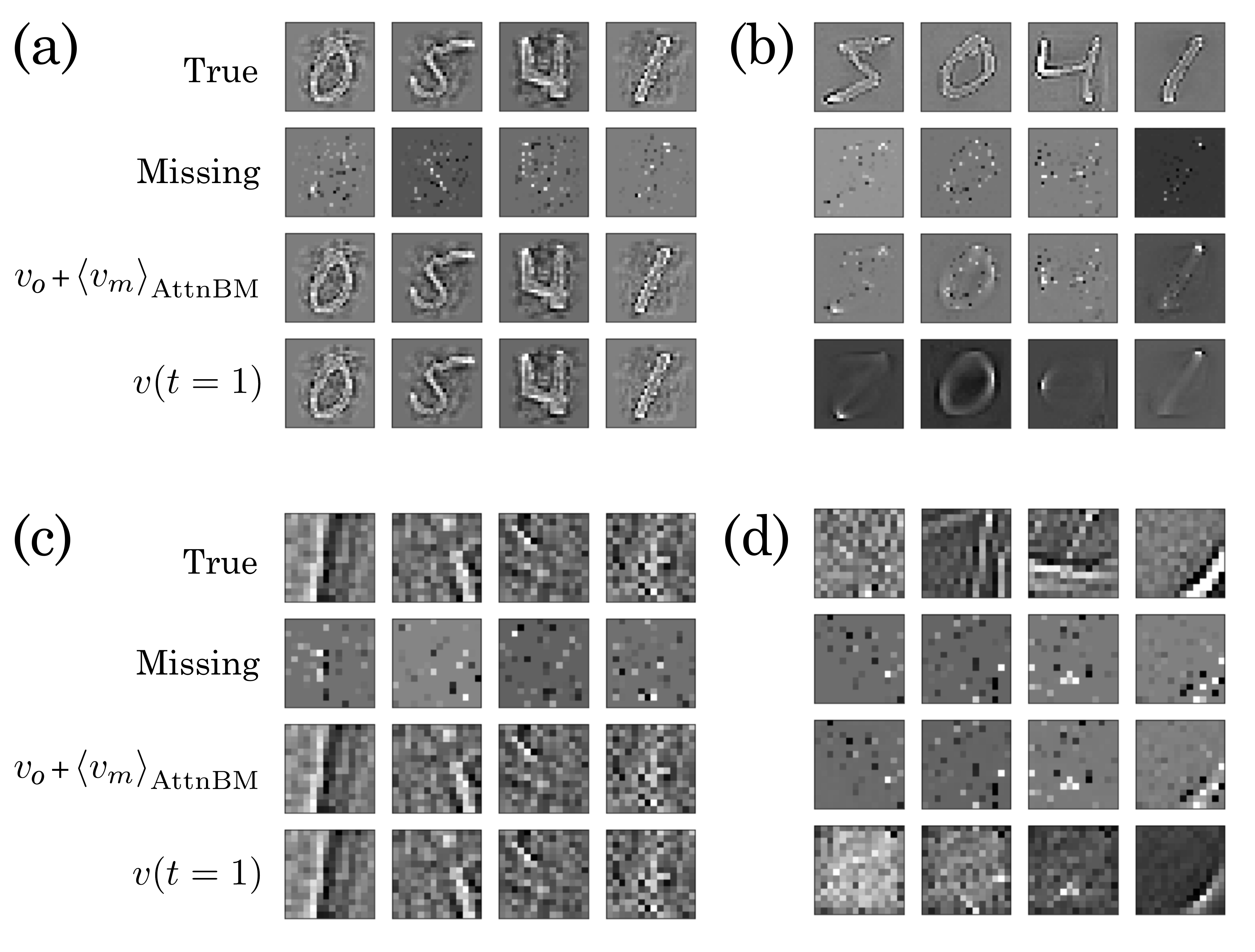}
    \caption{Image reconstruction on MNIST and the van Hateren natural images.
            (a) MNIST, P=200. (b) MNIST, P=50000. (c) van Hateren, P=200. (d) van Hateren, P=50000.}
    \label{fig2}
\end{figure}

\begin{figure}[t]
    \begin{minipage}[b]{0.5\linewidth}
        \centering
        \includegraphics[keepaspectratio, scale=0.45]{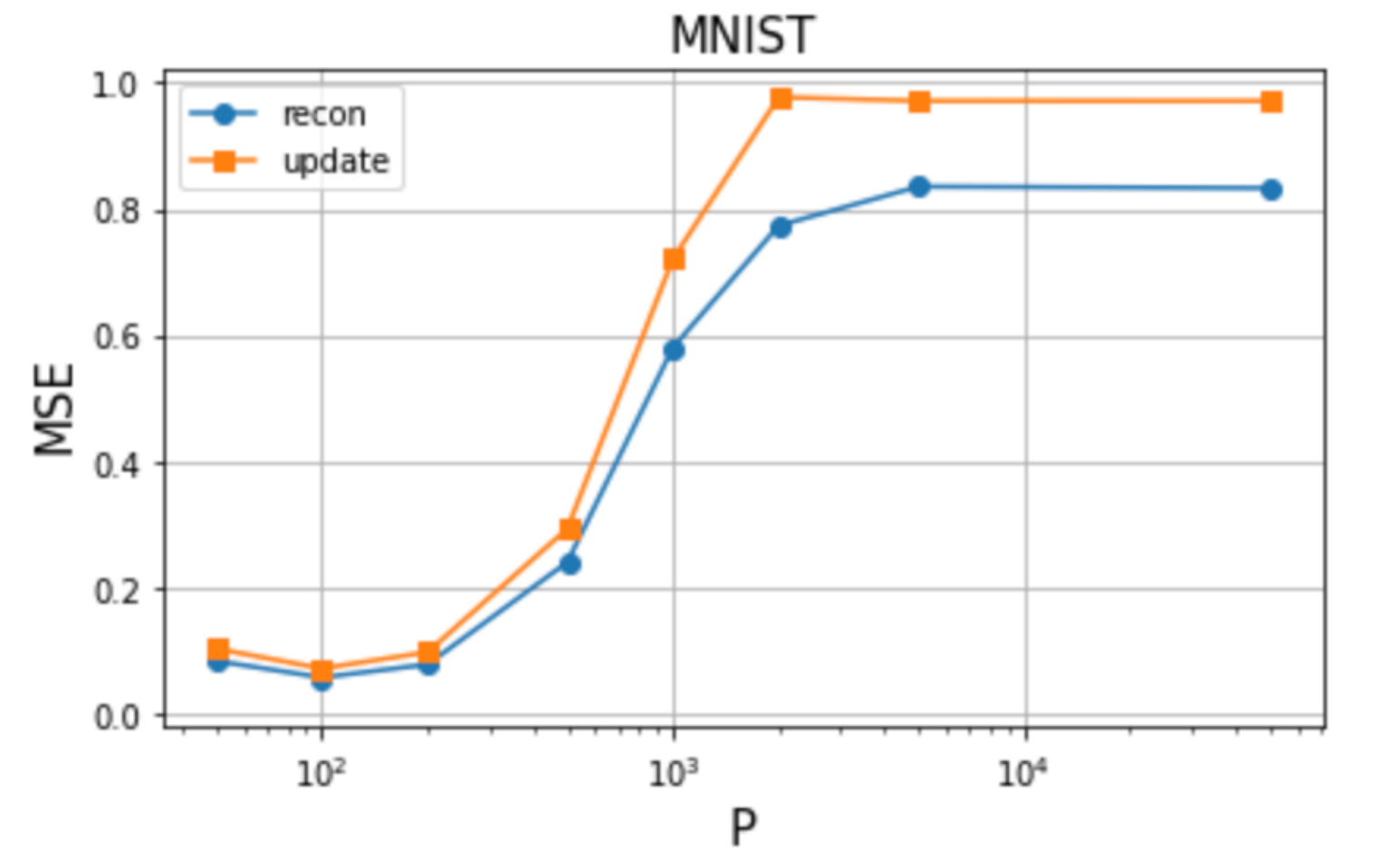}
        \label{fig:MSE-P_mnist}
    \end{minipage}
    \begin{minipage}[b]{0.5\linewidth}
        \centering
        \includegraphics[keepaspectratio, scale=0.45]{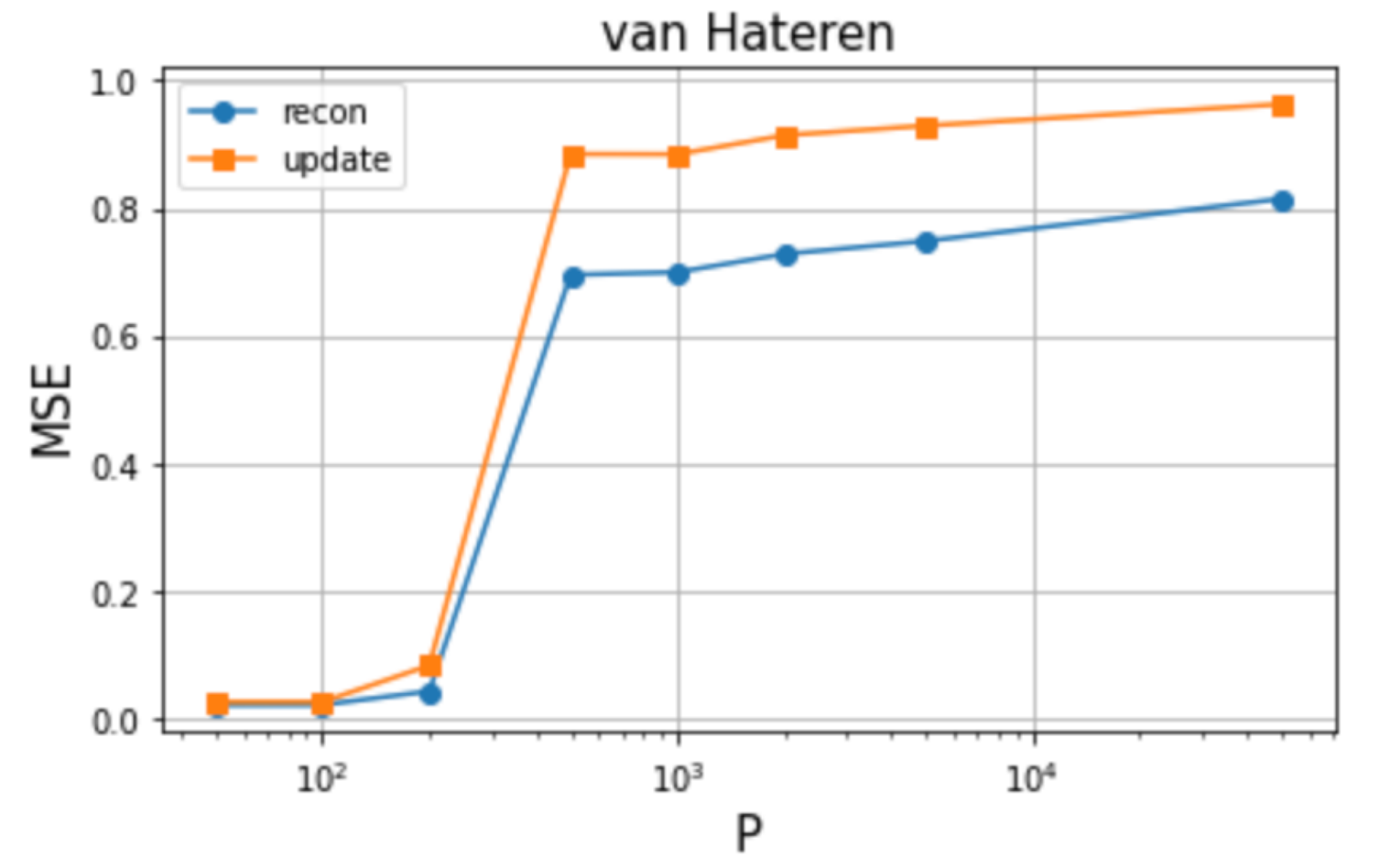}
        \label{fig:MSE-P_vanhateren}
    \end{minipage} 
    \caption{The MSE for the image reconstruction using Eq.~(\ref{eq:recon}) and the memory retrieval by the model-B update rule in Eq.~(\ref{eq:attn}), for the MNIST dataset (left) and the van Hateren natural images (right).}
    \label{fig:MSE-P}
\end{figure}

\fixme{%
Figure \ref{fig2} depicts the reconstruction of incomplete images using the trained \abm ~($N_h=900$).
The model was trained on $P=200,\, 50000$ training samples from MNIST or the van Hateren natural image dataset.
All the training data were pre-processed by ZCA whitening.
We computed the exact stochastic gradient descent (SGD) of the negative log-likelihood objective using Eq.~(\ref{eq27:1129}).
The mini-batch size was set to $5$ for $P=200$ and $50$ for $P=50000$.
The learning rate was set to $0.01$ for both, and no momentum and weight decay were used.
}%
The first line of the figures shows true training samples and the second one shows incomplete images,
where we randomly chose pixels and made them zero with a probability of 0.8.
The third line shows the reconstructed images that were obtained using the marginalized distribution of $p_{\text{\abm}}$.
Specifically, we consider a partition over the visible variable $\vv=[\vv_o; \vv_m]$,
where $\vv_o$ corresponds to observed variables and $\vv_m$ to missing ones.
The conditional distribution $p_{\text{\abm}}(\vv_m|\vv_o)$ is expressed as
\begin{multline}
    p_{\text{\abm}}(\vv_m|\vv_o)
        = (2\pi)^{-N_v'/2} \\
            \times \sum_\mu
            \softmax\left({\bm \xi}_{\mu,o}^{\top} \vv_o+ \frac{\|{\bm \xi}_{\mu,m}\|^2}{2}\right)
            \exp\left(-\frac{ \|\vv_m-{\bm \xi}_{\mu,m}\|^2}{2}\right),
    \label{eq:recon}
\end{multline}
where the weight vector is divided into ${\bm \xi}_\mu=[{\bm \xi}_{\mu,o};{\bm \xi}_{\mu,m}]$ and $N_v'$ denotes the dimension of $\vv_m$.
The missing variables could be reconstructed by averaging over $\vv_m$:
\begin{equation}
    \langle \vv_m \rangle_{\text{\abm}}
        = \sum_\mu {\bm \xi}_{\mu,m} \, \softmax\left({\bm \xi}_{\mu,o}^{\top} \vv_o + \frac{\|{\bm \xi}_{\mu,m}\|^2}{2}\right).
\end{equation}
\fixme{%
The third line presents this $\langle \vv_m \rangle_{\text{\abm}}$ for the missing part,
and it can be observed that the trained \abm ~reconstructed the true images effectively for the $P=200$ cases (Figs.~\ref{fig2}a,\ref{fig2}c).
In Figs.~\ref{fig2}b,\ref{fig2}d for $P=50000$, 
the model did not reconstruct the true images.
This is because the model did not memorize the each sample for the large sample size, but acquired local features of data, as will be seen below.
The fourth line displays the results of the memory retrieval only by a single update in the model-B Hopfield network.
}%
We substituted the trained weight obtained by the maximum-likelihood estimation of \abm ~into the update rule in Eq.~(\ref{eq:attn}).
Subsequently, we initialized the visible neurons using incomplete images and obtained $\vv(t)$.
We empirically confirmed that only a limited number of updates is sufficient to obtain the converged values.
\fixme{From Figs.~\ref{fig2}a,\ref{fig2}c}, we can see that the update rule could retrieve the true images only by a single update $\vv(t=1)$.
This is rational because \citep{ramsauer2021hopfield} analyzed the model B and found that it achieves rapid convergence to the embedded memories by a single update.
\com{%
    We also trained \abm ~for $P=\{50, 100, 200, 500, 1000, 2000, 5000, 50000\}$, with five runs for each.
    The mini-batch size was set to $5$ for $P=50, 100, 200$ and $50$ for the others.
    The rest of the experimental setups is taken exactly the same as above for all the $P$.
    After the training, we randomly picked 50 samples up from the training images and computed the mean squared error (MSE)
    between the inputs (true images) and the reconstructed images using the conditional distribution in Eq.~(\ref{eq:recon}) and the model-B update rule in Eq.~(\ref{eq:attn}), see Fig.~\ref{fig:MSE-P}.
    Figure \ref{fig:MSE-P} indicates that the models can efficiently reconstruct the true images for small $P$, and cannot for relatively large $P$.
    In Fig.~\ref{fig:MSE-P}, we only plotted the means of the results, and did not the error bars,
    since for large $P$ the models do neither memorize nor reconstruct the training samples and thus the standard deviations become quite large and meaningless.
}%

\begin{figure}[t]
    \centering
    \includegraphics[width=0.6\textwidth]{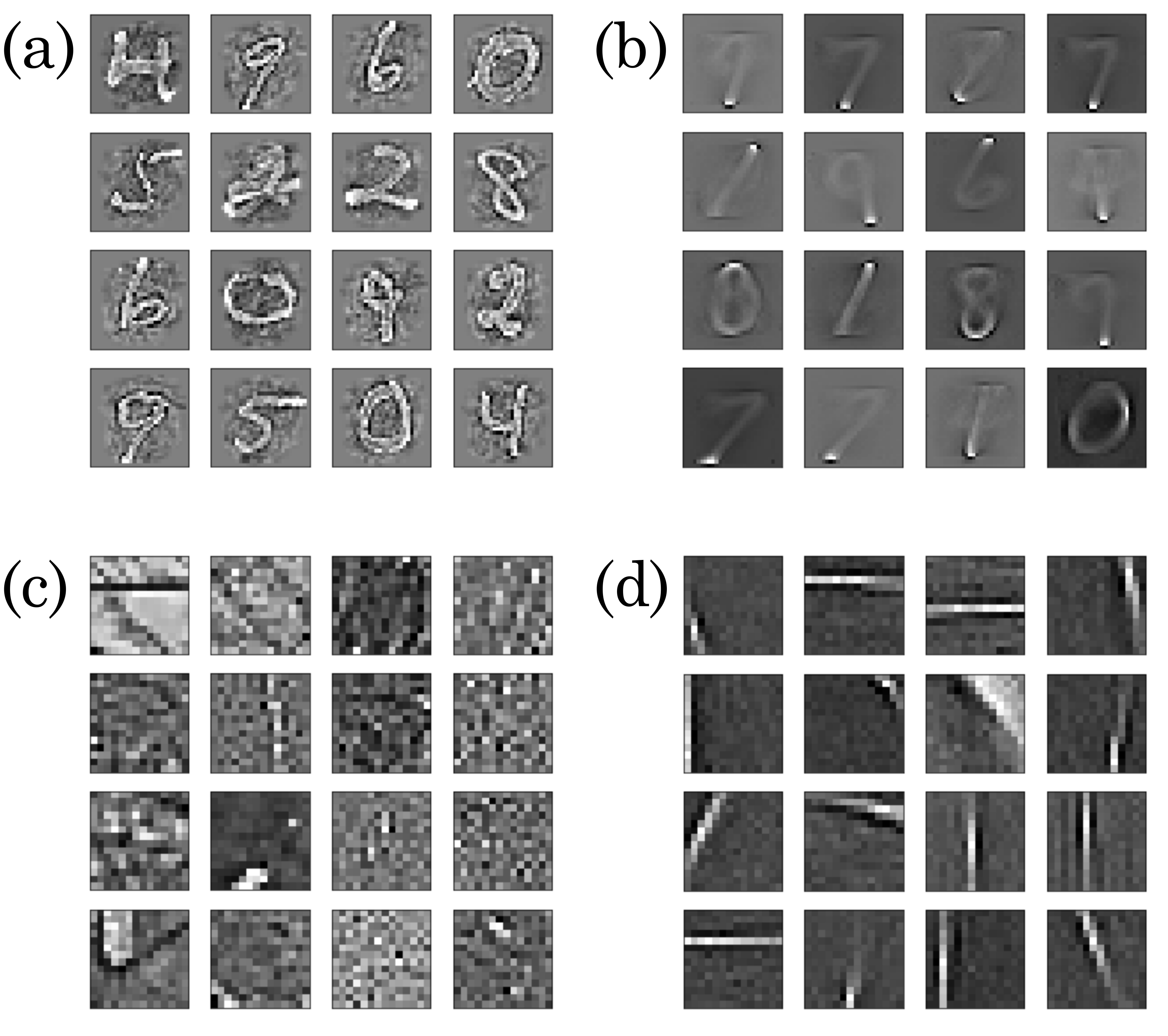}
    \caption{Visualization of filters obtained by \abm ~trained on MNIST and the natural images.
            (a) MNIST, P=200. (b) MNIST, P=50000. (c) van Hateren, P=200. (d) van Hateren, P=50000.}
    \label{fig3}
\end{figure}

\fixme{%
Figure \ref{fig3} shows the receptive fields (weight vectors ${\bm \xi}_\mu$ reshaped to $\sqrt{N_v}\times \sqrt{N_v}$ images)
of \abm ~obtained by the training on MNIST and the van Hateren natural images.
As noted above, for the $P=200$ cases the model memorized each sample in the weight vectors,
while the model extracted the local features of data for $P=50000$.
We can observe that the trained \abm ~obtained some edges of numbers for MNIST and the Gabor-like filters for the van Hateren natural images.
Such a local feature extraction is widely observed in unsupervised learning of various models including G--B RBM and autoencoders \citep{coates2011analysis}.
}%

\section{Stochastic description of Models A and C}
\label{sec:modelAC}

\subsection{Model-A Boltzmann Machine as an EFH}
\label{sec:modelA}

As discussed in \citep{krotov2021large},
the model A of the \kh ~is a general formulation including the classical Hopfield network
and dense associative memory models (DAMs) \citep{hopfield82,NIPS2016_eaae339c,demircigil2017model}.
This model is characterized by the energy function in Eq.~(\ref{eq:eng}) with two Lagrangians satisfying additivity:
\begin{equation}
    L_h(\vh) = \sum_\mu F(h_\mu),
    \quad
    L_v(\vv)= \sum_i |v_i|.
\end{equation}
We consider a general additive $L_v(\vv)=\sum_i G(v_i)$ below.
$F$ and $G$ are differentiable functions and provide the activation functions $f$ and $g$ by Eqs.~(\ref{eq6:1126}), respectively.

Let us consider the Boltzmann machine counterpart of Model A.
We find that this belongs to Exponential Family Harmoniums (EFHs) \citep{welling2004exponential} as follows.
An EFH is defined by
\begin{equation}
    p(\vv,\vh) = \exp
        \left[
            \sum_{i a} \theta_{i a} r_{i a}(v_i)
            +\sum_{j b} \lambda_{j b} s_{j b}(h_j)
            +\sum_{i j a b} J_{i a}^{j b} r_{i a}(v_i) s_{j b}(h_j)
        \right]/Z.
\end{equation}
One can see that the EFH is a general two-layer model including RBMs.
EFHs satisfy conditional independence properties; that is, $p(\vh|\vv) = \prod_i p(h_i|\vv)$ and $p(\vv|\vh) = \prod_i p(v_i|\vh)$,
which is desirable in contrastive divergence learning.
These conditional distributions are expressed as
\begin{align}
    \begin{aligned}
    &p(v_i | \vh )
        = \exp \left[\sum_a \hat{\theta}_{i a} r_{i a}(v_i) \right]/Z_{v_i|\vh},
        \quad
        \hat{\theta}_{i a}=\theta_{i a}+\sum_{j b} J_{i a}^{j b} s_{j b}(h_j), \\
    &p(h_j | \vv )
        = \exp \left[\sum_b \hat{\lambda}_{j b} s_{j b}(h_j)\right]/Z_{h_j|\vv},
        \quad
        \hat{\lambda}_{j b}=\lambda_{j b}+\sum_{i a} J_{i a}^{j b} r_{i a}(v_i),
    \end{aligned}
\end{align}
where the normalization factors $Z_{v_i|\vh}$ and $Z_{h_j|\vv}$ are relatively easy to evaluate because the conditional distributions are one-dimensional.
Here, we define the indices $a, b$ over $\{1,2\}$.
By setting
\begin{align}
    r_{i1}
        &= g_i(\vv),
        \quad
        r_{i2} = G(v_i)-v_i g_i(\vv), \\  
    s_{j1}
        &= f_j(\vh),
        \quad
        s_{j2} = F(h_j)-h_j f_j(\vh), \\
    \theta_{i1}
        &= \lambda_{j1} = 0,
        \quad
        \theta_{i2} = \lambda_{j2} = \beta, \\
    J_{ia}^{jb}
        &=\beta \xi_{ij} \ (a=b=1),
        \quad
        0 \ (\text{otherwise}),
\end{align}
one can see that the Boltzmann machine with the model-A energy, i.e., $p(\vv,\vh)=\exp(-\beta E_{\mathrm{A}}(\vv,\vh))/Z$, belongs to the EFHs.
We can easily apply contrastive divergence learning to them,
which implies that the Boltzmann machine counterpart of DAMs can be trained in the same manner as RBMs.
Note that we have not assumed the adiabatic limit (\ref{eq:hadia}) in this case.
If we substitute this limit instead of marginalizing $\vh$,
the obtained distribution $p(\vv)$ does not necessarily belong to the marginalized distribution of the EFH and may be hard to train.
Therefore, the use of the model with the joint energy $E_{\mathrm{A}}(\vv,\vh)$ is preferable from a trainability perspective.

It should be noted that Model B does not belong to the EFHs because the Lagrangian $L_h(\vh)$ in Eq.~(\ref{eq:Lattn}) has no additivity;
which means that the conditional independence of the distributions is not guaranteed
and learning with contrastive divergence is expected to be insufficient.
Fortunately, the model-B Boltzmann machine, i.e., \abm, has the tractable likelihood function for a positive integer $\beta$, which enables us to train the model very easily.
This implies that the tractability of the likelihood function does not necessarily corresponds to the class of EFHs.

\subsection{Remark on Model-C Boltzmann machine}

Model C, the final example discussed in \citep{krotov2021large}, is defined by the following Lagrangians:
\begin{equation}
    L_h(\vh) = \sum_\mu F(h_\mu),
    \quad
    L_v(\vv) = \sqrt{\sum_i v_i^2}.
\end{equation}
As $L_v(\vv)$ has no additivity, Model-C Boltzmann machine does not belong to EFHs.
Furthermore, it is unclear that the likelihood function becomes analytically tractable for a general $F$.
However, if we set $F$ to be the identity, $F(x)=x$, we have
\begin{align}
    E_{\mathrm{C}}
        &= -\sum_{\mu=1}^{N_h} F\left(\sum_{i=1}^{N_v} \xi_{\mu i} {v_i}/\|\vv\|\right)
        = -\frac{{\bm \eta}^\top \vv}{\sqrt{\sum_j v_j^2}},
\end{align}
where $\eta_i := \sum_\mu \xi_{\mu i}$ and note that we take the adiabatic limit in (\ref{eq:hadia}).
Therefore, its stochastic model is given by $p(\vv)= \exp(-\beta E_{\mathrm{C}}(\vv))/Z = \exp(\beta {\bm \eta}^\top \vv/\|\vv\|)/Z$,
which is the von Mises--Fisher distribution with the mean direction ${\bm \eta}/\|{\bm \eta} \|$ and concentration parameter $\|{\bm \eta}\|$.
\fixme{%
    The normalization factor is explicitly written as
\begin{align}
    Z = \frac{\|{\bm \eta}\|^{N_v/2 - 1}}{(2\pi)^{N_v/2} I_{N_v/2 - 1}(\|{\bm \eta}\|)},
\end{align}
where $I_\alpha$ is the Bessel function of the first kind.
}%
The above von Mises--Fisher case lacks the interpretation as a neural network and appears to be a toy example.
We leave this as an open problem to find a more general Model-C Boltzmann machine which we can train in some efficient ways.

\section{Conclusion}

In this letter, we have studied a family of Boltzmann machines given by Lagrangians and energy functions from modern Hopfield networks.
In particular, we have presented the novel \abm, which has a tractable likelihood function and gradient
for certain  special cases of the inverse temperature, and clarified that the special cases cover more general cases.
We also demonstrated that the Boltzmann machine of Model A belongs to EFHs.
This implies that the BM counterpart of dense associative memory models can also be trained in the same manner as RBMs.
The Boltzmann machine of Model C still lacks an efficient computation for training with the maximum likelihood, and further studies are desired.

A natural direction for further study is to take account of other choices of Lagrangians.
As the \kh ~can produce new energy-based associative memory models by replacing the Lagrangians,
the potential exists to generate new Boltzmann machine models that possess preferable properties for practical purposes.
The formulation in terms of Lagrangians and the associated energy functions may enable us
to explore Boltzmann machines with different architectural arrangements of neurons in a more transparent way.
We leave this aspect of study to future research.

\bibliography{references}
\bibliographystyle{apa}

\end{document}